\pdfoutput=1

\documentclass[11pt]{article}

\usepackage[final]{acl}

\usepackage{times}
\usepackage{latexsym}

\usepackage[T1]{fontenc}

\usepackage[utf8]{inputenc}

\usepackage{microtype}

\usepackage{inconsolata}

\usepackage{enumitem}
\usepackage{tabularx}
\usepackage{amsmath}
\usepackage{amsfonts}
\usepackage{amssymb}
\usepackage{tabu}
\usepackage{multirow}
\usepackage{multicol}
\usepackage{colortbl}
\usepackage{subcaption}
\usepackage{graphicx}
\usepackage{caption}
\usepackage{hhline}
\usepackage{booktabs}
\usepackage{tabularray}
\usepackage{hhline}

\newcommand{\highlight}[1]{{\textcolor{red}{\textbf{#1}}}}
\newcommand{\newtext}{}
\newcommand{\newstorytext}{}

\title{ROUGE-K: Do Your Summaries Have Keywords?}

\author{Sotaro Takeshita\textsuperscript{1}, Simone Paolo Ponzetto\textsuperscript{1}, Kai Eckert\textsuperscript{2} \\
  \textsuperscript{1}Data and Web Science Group, University of Mannheim, Germany \\
  \textsuperscript{2}Mannheim University of Applied Sciences, Mannheim, Germany \\ 
  \texttt{\{sotaro.takeshita, ponzetto\}@uni-mannheim.de} \\
  \texttt{k.eckert@hs-mannheim.de}}

\begin{document}
\maketitle
\begin{abstract}
Keywords, that is, content-relevant words in summaries play an important role in efficient information conveyance, making it critical to assess if system-generated summaries contain such informative words during evaluation.
However, existing evaluation metrics for extreme summarization models do not pay explicit attention to keywords in summaries, leaving developers ignorant of their presence.
To address this issue, we present a keyword-oriented evaluation metric, dubbed ROUGE-K, which provides a quantitative answer to the question of -- \textit{How well do summaries include keywords?}
Through the lens of this keyword-aware metric, we surprisingly find that a current strong baseline model often misses essential information in their summaries.
Our analysis reveals that human annotators indeed find the summaries with more keywords to be more relevant to the source documents. This is an important yet previously overlooked aspect in evaluating summarization systems.
Finally, to enhance keyword inclusion, we propose four approaches for incorporating word importance into a transformer-based model and experimentally show that it enables guiding models to include more keywords while keeping the overall quality.\footnote{Our code is released at \url{https://github.com/sobamchan/rougek}.}
\end{abstract}

\section{Introduction}
Summarization systems compress long documents into shorter ones to convey important information more effectively to readers \citep{rush-etal-2015-neural,chopra-etal-2016-abstractive}.
To convey all essential information correctly, it is crucial for summarization systems to include important, i.e., summary-relevant keywords.
In our analysis, human annotators find that summaries with more keywords, words that are relevant for the summary (see Section \ref{se:rouge-k}), capture important information better than the ones with fewer keywords.
However, existing evaluation metrics do not explicitly take such word importance into account.
\newstorytext{Table \ref{ta:rouge-problems} shows an example. Two commonly used metrics, namely ROUGE \citep{lin-2004-rouge} and BERTScore \citep{zhang-etal-2020-bertscore}, assign lower scores to the second hypothesis even though it contains an essential word that another summary misses. This discrepancy, namely that ROUGE assigns a lower score to a summary that contains more keywords and annotators find relevant, happens in 16.7\% of the cases in our analysis.}

\begin{table}[t]
    \footnotesize
    \centering
    \setlength{\tabcolsep}{4pt}
    \begin{tabularx}{\columnwidth}{| X | c | c |}
    \hline
    \multicolumn{3}{| l |}{\textbf{Reference}} \\ 
    \hline
    \multicolumn{3}{|p{\dimexpr \columnwidth-2\tabcolsep}|}{{A novel, hybrid deep learning approach provides the best solution to a limited-data problem (which is important to the conservation of the \highlight{\mbox{Hawaiian}} language)}}  \\
    \hline
     & \textbf{R-1} & \textbf{BS} \\ 
    \hline
    \multicolumn{1}{| l |}{\textbf{Hypothesis 1:}} & 27.45 & 0.8718 \\ 
    We propose two methods to solve the transliteration problem automatically, given that there were not enough data to train an end-to-end deep learning model. & & \\ 
    \hline
    \multicolumn{1}{| l |}{\textbf{Hypothesis 2:}} & 26.09 & 0.8692 \\ 
    We propose two methods to solve the \highlight{Hawaiian} orthography transliteration problem automatically using finite state transducers and a hybrid neural network. & & \\
    \hline
    \end{tabularx}
    \vspace{-0.5em}
    \caption{An example where ROUGE and BERTScore (BS) can lead to misinterpretations. Although the incorrect generation (not including the word ``Hawaiian'') in the first hypothesis is more critical than the one in the second summary (``neural network'' instead of ``deep learning'') to convey correct information, both metrics assign a higher score to the former summary.}
    \vspace{-0.5em}
    \label{ta:rouge-problems}
\end{table}

In this paper, to shed light on this problem, we propose ROUGE-K, an extension of ROUGE which considers only those n-grams in the summaries that match a set of pre-defined keywords.
We propose a simple heuristic that exploits the common structure of summarization datasets to extract keywords automatically, making it possible for our metric to scale in size and domain without additional annotation effort.
Correlation analysis reveals that there is only a weak strength of dependence between our new metric and existing ones as well as summary lengths, thus demonstrating that our approach can complement, rather than replace, previous metrics.
Through a manual evaluation, we find that human annotators show substantially higher agreement with \mbox{ROUGE-K} than with \mbox{ROUGE} and \mbox{BERTScore} on relevance, in other words, how well summaries include important information, which is one of four commonly assessed aspects in manual evaluations of summaries \citep{fabbri-etal-2021-summeval}. \newstorytext{This shows that while one still would use traditional ROUGE to assess the overall qualities, our metric can provide a better index for evaluating the relevance of summaries.}
As a showcase of this new metric, we evaluate both extractive \citep{liu-lapata-2019-text} and abstractive \citep{lewis-etal-2020-bart,dou-etal-2021-gsum,saito-etal-2020-abstractive} state-of-the-art models on two extreme summarization datasets from different domains, namely XSum \citep{narayan-etal-2018-dont} and SciTLDR \citep{cachola-etal-2020-tldr}, as well as a more traditional, non-extreme dataset, ScisummNet \citep{yasunaga_scisummnet_2019}. Besides news text (XSum), we choose summarization of scientific publications (ScisummNet and SciTLDR), since this is a domain where keyword inclusion within summaries plays a crucial role.  
Surprisingly, the results reveal that these strong baseline models often fail to include essential words in their summaries, and that ROUGE-K enables us to better distinguish systems' performance than alternative metrics. \newtext{We also apply our ROUGE-K to the evaluations of recent large language models (LLMs) and show how our metric better accounts for the powerful capabilities of LLM-based summarizers when compared to traditional ROUGE metrics.}
Finally, As a first attempt to address the limitations on summary keyword inclusion, we introduce four ways to incorporate a lightweight word importance feature into existing transformer-based models. Experiments show that our methods can guide models to include more keywords without any additional annotations and negative effects on overall summarization quality. Our contributions are the following ones:
\begin{itemize}[itemsep=0mm,leftmargin=3mm]
    \item We introduce \textbf{a new keyword-oriented evaluation metric, dubbed ROUGE-K}, which complements existing metrics by focusing on keywords.
    \item \textbf{We validate our metric}: a) against human judgments of summary relevance, b) by quantifying its correlation to existing metrics and summary lengths, and c) its ability to distinguish performance among different systems.
    \item Our experiments on three different datasets for summarization of scientific and news articles reveal that current \textbf{state-of-the-art models often fail to include important words} in summaries.
    \item We present experiments with \textbf{four approaches to incorporate word-importance scores} into BART and show that it can help to improve keyword inclusion without hurting the overall summarization qualities.
\end{itemize}
\section{The need for another kind of ROUGE}
\label{se:problems}

ROUGE \citep{lin-2004-rouge} is a long-running \textit{de facto} standard evaluation metric for summarization systems. It is very popular due to its high correlation with human evaluation while keeping its simplicity and interpretability. However, several works report on its limitations \citep{akter-etal-2022-revisiting,fabbri-etal-2021-summeval}, one of which is that it only takes the word surface into account and disregards semantics \citep{ng-abrecht-2015-better}.
Because it considers all n-gram matches to be equally important, ROUGE fails to detect salient words that underpin a summary's quality.

As an example, Table \ref{ta:rouge-problems} shows two generated summaries of the same article from the SciTLDR dataset \citep{cachola-etal-2020-tldr} as well as their scores computed by ROUGE and BERTScore \citep{zhang-etal-2020-bertscore}, a pre-training language model based metric.
Both metrics assign a higher score to the first summary even though it misses an important keyword that the second summary contains. In the case of ROUGE, this is because it favors the longer but nonessential n-gram overlaps in the first summary.
This limitation of evaluation metrics can mislead the development of summarization systems towards including more of longer but less important words in summaries than truly essential keywords.
When multiple reference summaries are available, ROUGE can assign higher scores to words potentially more important than others by counting n-grams that appear several times across references, which indirectly considers word importance.
However, most commonly used datasets contain only one reference summary \citep{hermann-etal-2015-teaching,narayan-etal-2018-dont}.
In addition, because of its implicit nature, when a generated summary has a different textual style (even if the semantics of the summary did not change) from its reference summary, the ROUGE score can easily deflate.
\section{ROUGE-K}
\label{se:rouge-k}
We present ROUGE-K, an extension of ROUGE that exclusively focuses on essential words in summaries.
Its core idea is simple: ROUGE-K assesses the proportion of keywords from the reference summary that are included in the candidate summary. We compute ROUGE-K as:
\begin{equation*}
    \text{R-K} = \frac{\textrm{Count}(\textrm{kws} \cap \textrm{n-grams})}{\textrm{Count}(\textrm{kws})}
\end{equation*}
where $\textrm{kws}$ is a set of pre-defined keywords and $\textrm{n-grams}$ is a target hypothesis.
This provides a direct understanding of how well system summaries contain essential pieces of information.
ROUGE-K is essentially a recall-oriented metric since it computes \textit{coverage} of keywords. While it is possible to complement this formula with another one to compute precision, this would give the proportion of keywords in the candidate summary. However, this metric would indicate how good the system is at extracting keywords, not its summarization capabilities, i.e., one could have a summary consisting only of keywords but only marginally overlapping with the reference summary.

\paragraph{Keyword extraction.}
\label{se:keyword-extraction}
\begin{table}[t]
\small
\centering
\begin{tabularx}{\columnwidth}{| X |}
\multicolumn{1}{c}{SciTLDR \citep{cachola-etal-2020-tldr}}\\
\hline
We show that autoregressive models can generate \highlight{high} \highlight{fidelity images}. \\ 
\hline
We introduce a new inductive bias that \highlight{integrates} \highlight{tree structures} in \highlight{recurrent} neural networks. \\
\hline
\multicolumn{1}{c}{} \\
\multicolumn{1}{c}{XSum \citep{narayan-etal-2018-dont}} \\
\hline
\highlight{Opec}, the \highlight{oil} producers' group is \highlight{back} in the driving seat. \\
\hline
\highlight{Lenovo and Acer} have both unveiled \highlight{smartphones} with much larger than normal \highlight{batteries}.\\
\hline
\end{tabularx}
\caption{Sample reference summaries with highlights on keywords extracted by our heuristic.}
\label{ta:keyword-samples}
\end{table}
An essential prerequisite of ROUGE-K is the availability of keywords. Ideally, we would like these keywords to be available for \textit{any} summarization corpus to enable the wide applicability of our metric. A solution is thus to extract keywords from reference summaries heuristically.
\citet{nan-etal-2021-entity}, for instance, use words detected by a named entity recognition (NER) model to evaluate entity-level factual consistency in summaries.
However, (1) not all keywords are named entities, (2) NER models accurate enough to be used for evaluation are not available for all domains (e.g., scholarly documents), (3) the accuracy of NER models for documents in summarization datasets is unknown.

In this paper, we present a simple and interpretable way to extract keywords.
We define keywords as \emph{the n-grams used in multiple reference summaries}, assuming that words used in multiple human-written summaries for the same document repeatedly should be included in system summaries as well.
First, we tokenize and lowercase the reference summaries, extract n-grams, and then remove stopwords from the extracted n-grams.
Next, we compare n-grams from multiple references and extract those that appear in multiple references. To capture multi-word keywords, the extraction process starts from 10-grams to unigrams.
\emph{When there is only one reference summary available, the corresponding title is used as a proxy reference} which is known to contain key information \citep{koto-etal-2022-lipkey, cachola-etal-2020-tldr}. Table \ref{ta:keyword-samples} shows examples of keywords extracted by our heuristic.
We benchmark our heuristic against TF-IDF \citep{salton-etal-1988-term} and TextRank \citep{mihalcea-tarau-2004-textrank}. To this end, we take the first 100 samples of the SciTLDR development data and for each summary, we extract the same amount of keywords as the one from our method (i.e., we keep the recall level fixed). We then quantify for each method a) the average number of wrong keywords per summary and b) the overall false discovery rate FDR (for both, lower is better) -- our hunch is that for humans, it is easier to judge whether something is not a keyword, as opposed to being one. In both cases, our heuristic achieves the best performance: 0.64 vs.\ 0.85 and 0.94 on average wrong extractions per summary and 0.13 vs.\ 0.16 and 0.21 FDR when compared against TF-IDF and TextRank, respectively.

\paragraph{Agreement with human judgments.}
\begin{table}[t]
    \small
    \centering
    \setlength{\tabcolsep}{5.9pt}
    \begin{tabular}{lccccc}
        \toprule
         & \textbf{R-1} & \textbf{R-2} & \textbf{R-L} & \textbf{BS} & \textbf{R-K} \\ 
        \midrule
        SciTLDR & 61.11 & 58.89 & 60.00 & 57.78 & 72.22 \\
        XSum & 63.73 & 59.80 & 56.86 & 62.75 & 70.59 \\
        \bottomrule
    \end{tabular}
    \caption{\newstorytext{Agreement ratios (\%) of each metric and human annotator on summary relevance, computed as the proportion of documents for which a given metric gives the highest score to the summary judged as most relevant from humans.}}
    \label{ta:humaneval}
\end{table}
We now perform a manual evaluation to test how well \mbox{ROUGE-K} aligns with human judgements on rating the relevance of summaries (we follow \citet{fabbri-etal-2021-summeval} and define `relevance' as the \textit{selection of important content from the source}. \newtext{We focus for manual evaluation on relevance only (as opposed to, e.g., \citet{fabbri-etal-2021-summeval} considering three other aspects) because the purpose of ROUGE-K is to quantify how well summaries include essential words, and thus preserve important, i.e., relevant content, as opposed to, e.g., ROUGE taking into account style aspects.

Our dataset consists of pairs of summaries generated using different instances of the same model (BART), trained on each of SciTLDR and XSum with different random seeds. To avoid ties, we select a sample of 92 and 100, respectively from SciTLDR and XSum, summary pairs where the two models assign a different ROUGE-K score to each summary. We then ask four annotators from our CS graduate course to compare the summaries and rank them (i.e., label the best one among the two). We finally compute how often each evaluation metric assigns higher scores to the summaries preferred by the annotators.
Results are shown in Table \ref{ta:humaneval}.
In line with \citet{fabbri-etal-2021-summeval}, R-1 shows higher agreement than R-2 and R-L, and BERTScore marks a marginally lower score than ROUGE-1.
Finally, \mbox{ROUGE-K} shows much higher agreement, indicating its strong ability to detect human-preferable summarization models.

\paragraph{Benchmarking BART with ROUGE-K.}
\begingroup
\begin{table*}[t]
\small
\centering
\setlength{\tabcolsep}{3.5pt}
\begin{tabular}{l|c|c|c|c|c|c|c} 
\hline
 & \multicolumn{2}{c|}{\textbf{Documents}} & \multicolumn{3}{c|}{\textbf{Summaries}} & \multicolumn{2}{c}{\textbf{Extracted keywords}} \\ 
\hline
\textbf{Dataset} & \multicolumn{1}{c|}{\begin{tabular}[c]{@{}c@{}}\textbf{\# documents}\\\textbf{(train/val/test)}\end{tabular}} & \multicolumn{1}{c|}{\begin{tabular}[c]{@{}c@{}}\textbf{\# words}\\\textbf{per doc}\\\textbf{on avg.}\end{tabular}} & \multicolumn{1}{c|}{\begin{tabular}[c]{@{}c@{}}\textbf{\# words}\\\textbf{per summary}\\\textbf{on avg.}\end{tabular}} & \multicolumn{1}{c|}{\begin{tabular}[c]{@{}c@{}}\textbf{compress.}\\\textbf{ratio}\end{tabular}} & \multicolumn{1}{c|}{\begin{tabular}[c]{@{}c@{}}\textbf{\# references}\\\textbf{on avg.}\\\textbf{(train/val/test)}\end{tabular}} & \multicolumn{1}{c|}{\begin{tabular}[c]{@{}c@{}}\# \textbf{keywords}\\\textbf{on avg.}\\\textbf{(train/val/test)}\end{tabular}} & \multicolumn{1}{c}{\begin{tabular}[c]{@{}c@{}}\textbf{average}\\\textbf{lengths}\\\textbf{(train/val/test)}\end{tabular}} \\ 
\hline
SciTLDR & 1,992 / 619 / 618 & 5,000 & \hspace{0.55em}21.00 & 238.10 & 2.0 / 3.3 / 4.2 & 1.9 / 4.2 / 5.2 & 1.7 / 1.5 / 1.5 \\ 
\hline
XSum & 204K / 11K / 11K & \hspace{0.7em}431 & \hspace{0.55em}23.26 & \hspace{0.55em}18.53 & 2.0 / 2.0 / 2.0 & 2.9 / 2.9 / 2.9 & 1.5 / 1.5 / 1.5 \\
\hline
ScisummNet & 750 / 92 / 91 & 4,700 & 167.49 & \hspace{0.55em}28.06 & 2.0 / 2.0 / 2.0 & 2.8 / 3.0 / 2.6 & 1.7 / 1.6 / 1.6 \\
\hline
\end{tabular}
\caption{Statistic of datasets and extracted keywords.}
\label{ta:datasets}
\end{table*}
\endgroup

\begin{table}[t]
    \small
    \centering
    \setlength{\tabcolsep}{9pt}
    \begin{tabular}{lrrrr}
    \toprule
        & \multicolumn{1}{c}{\textbf{R-1}} & \multicolumn{1}{c}{\textbf{R-2}} & \multicolumn{1}{c}{\textbf{R-L}} & \multicolumn{1}{c}{\textbf{R-K}} \\ 
        \midrule
        SciTLDR & 43.93 & 22.31 & 36.58 & 41.36 \\
        XSum & 44.43 & 21.00 & 35.94 & 56.14 \\
        \midrule
        ScisummNet & 50.75 & 47.80 & 49.73 & 68.95 \\
        \bottomrule
    \end{tabular}
    \caption{BART performance evaluated by ROUGE-1/-2/-L and our ROUGE-K.}
    \label{ta:baseline-results}
\end{table}

As a showcase of ROUGE-K, we evaluate BART \citep{lewis-etal-2020-bart}, a strong transformer-based generative language model on three different datasets: \mbox{SciTLDR} \citep{cachola-etal-2020-tldr}, XSum \citep{narayan-etal-2018-dont} and ScisummNet \citep{yasunaga_scisummnet_2019}. These cover different summarization tasks -- i.e., extreme (\mbox{SciTLDR}, \mbox{XSum}) vs.\ non-extreme (\mbox{ScisummNet}) -- as well as different domains -- i.e., scholarly documents (\mbox{SciTLDR}, \mbox{ScisummNet}) vs.\ news (\mbox{XSum}). Datasets details are shown in Table \ref{ta:datasets}.
BART models are fine-tuned on the training set and early stopping is performed using validation data, and finally evaluated on the test split using the traditional ROUGE metrics and our ROUGE-K.
Table \ref{ta:baseline-results} shows the results.
Each score is an average of ten and three different random seeds, respectively, for SciTLDR and XSum/ScisummNet (a larger number of seeds is used for SciTLDR to obtain stable scores on its relatively small test dataset).
Although one would consider the scores achieved by BART on ROUGE-1/-2/-L to be high, it only reaches 41.36\% and 56.14\% on ROUGE-K. In other words, a strong baseline model fails to include half of the essential n-grams in its summaries, unveiling a critical limitation previously missed by standard metrics.

\paragraph{Correlation with summary lengths.}
\begin{table}[t]
    \small
    \centering
    \setlength{\tabcolsep}{8.3pt}
    \begin{tabular}{lrrrr}
        \toprule
        & \multicolumn{1}{c}{\textbf{R-1}} & \multicolumn{1}{c}{\textbf{R-2}} & \multicolumn{1}{c}{\textbf{R-L}} & \multicolumn{1}{c}{\textbf{R-K}} \\ 
        \midrule
        SciTLDR & -0.102 & -0.070 & -0.154 & 0.167 \\
        XSum & -0.003 & -0.037 & -0.075 & 0.057 \\
        \midrule
        ScisummNet & 0.356 & 0.435 & 0.392 & 0.402 \\
        \bottomrule
    \end{tabular}
    \caption{Pearson Correlation between the number of words in summaries and evaluation metrics.}
    \label{ta:length}
\end{table}

Since ROUGE-K is recall-oriented, it can potentially favor longer summaries, i.e., as suggested by the overall absolute scores obtained by BART in Table \ref{ta:baseline-results} on non-extreme summarization with ScisummNet data. To quantify this, we compute Pearson correlations between the number of words in summaries generated by BART and different evaluation scores. As Table \ref{ta:length} shows, ROUGE-K scores have marginally higher correlations with summary lengths than other ROUGE (F1) metrics, although they all are relatively weak, ranging from -0.07 to 0.17 on SciTLDR and even lower for XSum. These results are different from those from \citet{sun_how_2019}, arguably because SciTLDR and XSum are extreme summarization datasets. On non-extreme summarization (ScisummNet), the results align instead with previous findings. However, we observe the same level of moderate correlation with the summary length between vanilla ROUGE and ROUGE-K.

\paragraph{Correlation with existing metrics.}
\begin{table}[t]
    \small
    \centering
    \setlength{\tabcolsep}{9.8pt}
    \begin{tabular}{lrrr}
    \toprule
        & \multicolumn{1}{c}{\textbf{R-1 (avg)}} & \multicolumn{1}{c}{\textbf{R-1 (max)}} & \multicolumn{1}{c}{\textbf{BS}} \\ 
        \midrule
        SciTLDR & 0.510 & 0.434 & 0.383 \\
        XSum & 0.318 & 0.318 & 0.237 \\ 
        \midrule
        ScisummNet & 0.288 & 0.288 & 0.413 \\
        \bottomrule
    \end{tabular}
    \caption{Pearson Correlation between ROUGE-K and ROUGE-1 average, ROUGE-1 max and BERTScore.}
    \label{ta:corr}
\end{table}

To better understand the relationship between \mbox{ROUGE-K} and other existing metrics, we perform an additional correlation analysis (Table \ref{ta:corr}).
R-1 (avg) computes a \mbox{R-1} for each reference given a sample and takes the average while R-1 (max) takes only the largest score. R-1 (avg) and R-1 (max) are the same for XSum and ScisummNet because there is only one reference summary in this dataset.
The results indicate only a moderate strength of association between ROUGE-K and existing metrics, thus providing evidence that our metric can partially complement other metrics.
\section{Importance-guided summarization}
\label{se:new-models}
\begin{figure}[tb]
    \centering
    \includegraphics[width=0.8\columnwidth]{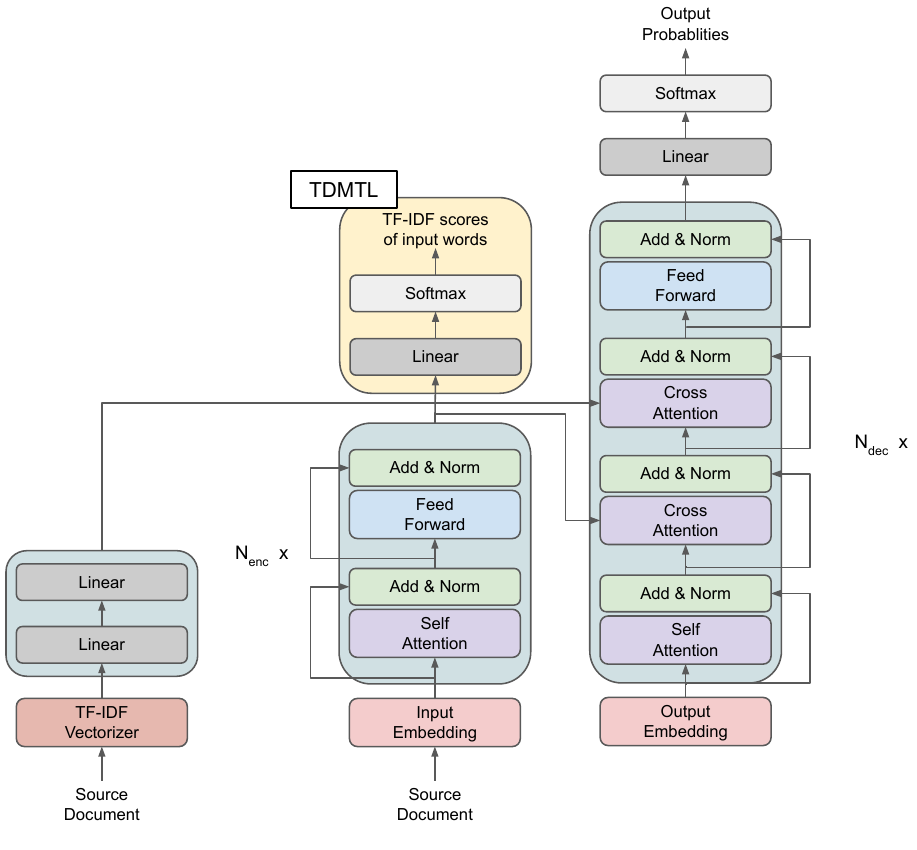}
    \caption{Overview of our TDSum model.}
    \label{fig:tdsum}
\end{figure}
We next propose four ways to incorporate a soft guiding signal into BART to enforce the inclusion of keywords into the generated summaries.

\paragraph{Re-weighted encoding (RwEnc).}
The first approach is to modify the representations within the model with TF-IDF scores.
\begin{equation*}
    \textrm{attention matrix} = \textrm{softmax}(\frac{QK^{\intercal}}{\sqrt{d_k}} + T) V
\end{equation*}
The first term within the softmax function is from the original transformer \citep{vaswani_attention_2017} where Q and K are query and key matrices respectively, and we introduce the second term $T$ which is a matrix of TF-IDF scores over the input text.
This enhances the model to propagate higher values for the important words to the upper layers. We apply this modification to the 0-, 4-, 8-th encoder layers, empirically selected on the dev data.

\paragraph{Re-weighted generation (RwGen).}
The second solution operates in the token selection phase.
At each generation step, BART computes a probabilistic distribution over its vocabulary for the next token to produce. We modify this distribution by summing TF-IDF scores so that the words with higher scores are more likely to be selected:
\begin{equation*}
\begin{aligned}
    \textrm{score}(y_{w_t}|w_{<t}, X, T)' &=\\
    (1 - \lambda) * \textrm{score}&(y_{w_t}|w_{<t}, X) + \lambda * T
\end{aligned}
\end{equation*}
where $\textrm{score}$ is a fine-tuned BART that takes previously generated words ($w_{<t}$) and the source document ($X$), and predicts scores which are further transformed to the probability for the next token at the time step $t$ by a softmax function.
We introduce the second term ($T$) which is a vector filled with TF-IDF values for the source document. $\lambda$ is a hyperparameter with which we control how much we shift the distribution from vanilla BART.

\paragraph{Multi-Task Learning with TF-IDF (TDMTL).}
Another solution is to modify the objective function to ask the model to predict TF-IDF scores in parallel with generating summaries. For this, we compute the mean squared error as loss for TF-IDF score prediction $L_{tfidf}$  and the standard cross entropy loss for summarization $L_{sum}$. The final loss we minimize is the linear interpolation of the two task-specific losses: $(1-\lambda) L_{sum} + \lambda L_{tfidf}$.

\paragraph{Injecting TF-IDF into the decoder (TDSum).}
Our last approach is inspired by \citet{dou-etal-2021-gsum}. Since their approach requires an explicit guidance signal (e.g., keywords), it uses additional models for keyword extraction leading to a drastic increase in computational costs.
Instead, we propose to use light-weight \mbox{TF-IDF} scores as shown in Figure \ref{fig:tdsum}.
TDSum equips two linear layers to process \mbox{TF-IDF} scores for words in input documents and uses resulting word importance features in newly introduced cross-attention layers in each decoder layer to guide the model towards keyword-oriented summary generation.
We train this model with the aforementioned TDMTL loss.
\section{Experiments and results}
\label{se:experiments}

\begin{table*}
\small
\centering
\setlength{\tabcolsep}{5.9pt}
\begin{tabular}{lrrrrrrrrrrrrrr}
\toprule
\multirow{2}{*}{\textbf{Model}} & \multicolumn{4}{c}{\textbf{SciTLDR}} & \multicolumn{1}{c}{} & \multicolumn{4}{c}{\textbf{XSum}}  & \multicolumn{1}{c}{} & \multicolumn{4}{c}{\textbf{ScisummNet}}\\ 
\cmidrule{2-5}\cmidrule{7-10}\cmidrule{12-15}
         & \multicolumn{1}{c}{\textbf{R-1}} & \multicolumn{1}{c}{\textbf{R-2}} & \multicolumn{1}{c}{\textbf{R-L}} & \multicolumn{1}{c}{\textbf{R-K}}  &  & \multicolumn{1}{c}{\textbf{R-1}} & \multicolumn{1}{c}{\textbf{R-2}} & \multicolumn{1}{c}{\textbf{R-L}} & \multicolumn{1}{c}{\textbf{R-K}}  & & \multicolumn{1}{c}{\textbf{R-1}} & \multicolumn{1}{c}{\textbf{R-2}} & \multicolumn{1}{c}{\textbf{R-L}} & \multicolumn{1}{c}{\textbf{R-K}}  \\ 
\midrule
BART & 43.93 & 22.31 & 36.58 & 41.36  &  & 44.43 & 21.00 & 35.94 & \textbf{56.14} & & 50.75 & 47.80 & 49.73 & 68.95  \\ 
GSum & 43.65 & 22.09 & 36.50 & 41.00  &  &  43.86 & 20.47 & 35.60 & 52.85 & & 24.37 & 21.11 & 23.35 & 43.36  \\
MTL & 43.82 & 22.24 & 36.29 & 42.83  &  & 44.50 & \textbf{21.05} & \textbf{36.10} & 56.06 & & 50.75 & 47.81 & 49.73 & 68.81  \\ 
PreSumm & 30.43 & 11.36 & 24.08 & 25.06 &  & 22.16 & 4.13 & 15.91 & 24.67 & & \textbf{60.58} & 49.15 & 46.22 & 68.85 \\ 
\midrule
Std (1--4) & 5.79 & 4.70 & 5.36 & \textbf{7.25} &  & 9.57 & 7.24 & 8.65 & \textbf{13.12} &  & \textbf{13.45} & 11.77 & 11.01 & 11.05 \\
\midrule
RwEnc & 43.98 & 22.39 & 36.68 & 41.03  &  &  44.42 & 20.93 & 36.07 & 55.58 & & 50.75 & \textbf{47.92} & \textbf{49.89} & 69.40  \\ 
RwGen & 43.96 & 22.35 & 36.59 & 41.60  &  & \textbf{44.57} & 21.04 & 36.09 & 56.03 & & 50.38 & 47.78 & 49.54 & 69.19 \\ 
TDMTL & \textbf{44.08} & \textbf{22.48} & \textbf{36.75} & 41.85  &  &  44.50 & \textbf{21.05} & \textbf{36.10} & 56.06 & & 50.64 & 47.75 & 49.67 & \textbf{69.56}  \\ 
TDSum & 43.55 & 21.74 & 35.82 & \underline{\textbf{43.04}}  &  & 44.13 & 20.95 & 35.57 & 55.39 & & 50.50 & 47.63 & 49.55 & 69.43  \\
\midrule
Std (all) & 4.44 & 3.60 & 4.10 & \textbf{5.59} &  & 7.34 & 5.56 & 6.62 & \textbf{10.23} & & \textbf{9.72} & 8.90 & 8.62 & 8.54 \\
\bottomrule
\end{tabular}
\caption{Results on SciTLDR, XSum, and ScisummNet. Best results per metric are \textbf{bolded}. Scores with \underline{underline} indicate that they significantly outperform all baseline models. We test for statistical significance using the Wilcoxon signed-rank test with $\alpha = 0.05$ \citep{dror-etal-2018-hitchhikers}.}
\label{ta:results}
\end{table*}

\subsection{Experimental setup}

\paragraph{Datasets.}
\label{se:datasets}

We experiment on different domains and summarization tasks using SciTLDR \citep{cachola-etal-2020-tldr}, XSum \citep{narayan-etal-2018-dont} and ScisummNet \citep{yasunaga_scisummnet_2019}

\paragraph{Baselines.}
\label{se:baseline_models}
We compare our models with three abstractive and one extractive summarizer:
\begin{itemize}[itemsep=0mm,leftmargin=2.7mm]
\item \textbf{BART} \citep{lewis-etal-2020-bart} is a transformer-based generative language model, pre-trained with denoising objective function.
\item \textbf{GSum} \citep{dou-etal-2021-gsum} is an extension of BART with additional parameters for processing textual guidance signals. Here, we input overlapping keywords, extracted as explained in Section \ref{se:keyword-extraction}.
\item \textbf{MTL} \citep{saito-etal-2020-abstractive} performs multitask training to predict keywords in source documents in addition to the summarization objective (we use our extracted keywords from Section \ref{se:keyword-extraction}).
\item \textbf{PreSumm} \citep{liu-lapata-2019-text} is an extractive summarization model based on BERT \citep{devlin-etal-2019-bert}.
\end{itemize}

\paragraph{Hyperparameter tuning.} We perform a grid search for each dataset and model using the development data and ROUGE-1 as a reference.
We test for learning rate $\in \{1e-05, 2e-05, 3e-05\}$, gradient accumulation $\in \{4, 8\}$, number of beam search $\in \{2, 3\}$ and repetition penalty rate $\in \{0.8, 1.0\}$. 
We also explore $\lambda \in \{0.1, 0.2, 0.3\}$ for MTL and TDMTL and $\lambda \in \{30, 50\}$ for RwGen.
During hyperparameter search, we use one random seed.
The final reported results on the test data are the averaged performance over models fine-tuned with different random seeds. We use ten seeds for SciTLDR and ScisummNet and three for XSum.
Our experiments are performed on RTX A6000 and utilize the implementation by \citet{deutsch-roth-2020-sacrerouge} to compute ROUGE-1/2/L.

\subsection{Results and discussion}
We organize the discussion of our results around the following research questions:

\begin{itemize}[itemsep=0mm,leftmargin=3mm]
    \item \textbf{RQ1:} Can models that incorporate TF-IDF scores increase the number of keywords in the summaries without degrading ROUGE scores?
    \item \textbf{RQ2:} Which kinds of keywords do models find hard to include in summaries?
\end{itemize}

\paragraph{RQ1: TF-IDF as guidance.}
We present our main results in Table \ref{ta:results}. On the SciTLDR dataset, BART marginally outperforms two other baseline models on ROUGE-1/2/L. However, MTL performs the best on the ROUGE-K metric, thus showing its effectiveness of explicit training guidance.
As reported by previous works \citep{cachola-etal-2020-tldr,narayan-etal-2018-dont}, an extractive model considerably underperforms abstractive models on all the metrics in extreme summarization since it suffers from merging information across multiple input sentences into its outputs. Because keywords are also scattered over multiple sentences, it fails to include most keywords.
Three out of four of our newly introduced TF-IDF-equipped models outperform vanilla BART on keyword inclusion, and TDSum significantly outperforms all the baselines on ROUGE-K while keeping its ROUGE scores on par with BART.
TDMTL follows the same training procedure as MTL and learns to predict TF-IDF instead of keywords. While results still improve over BART, our results show that using hard signals (i.e., keywords) is preferable.
RwGen is simple and fast to train, yet it includes more keywords than BART.
On XSum, BART outperforms other baseline models on ROUGE-K. However, the MTL model exceeds other traditional ROUGE metrics showing that our metric can shed light on an aspect that other metrics cannot capture. Among our proposed methods, TDMTL performs well akin to the results on SciTLDR, outperforming BART on traditional ROUGE metrics, although the BART model still outperforms models with TF-IDF extensions on ROUGE-K.

We see similar trends for non-extreme summarization on ScisummNet, where our proposed models are on par (sometimes outperform) with baseline models on ROUGE metrics. All proposed methods outperform all baseline models on ROUGE-K, indicating that, even for longer summaries, TF-IDF can enhance keyword inclusion. One significant difference is that the extractive model (PreSumm) performs better than abstractive models on ROUGE-1/2. We speculate this is due to much longer output summaries (181.88 words on average for PreSumm vs.\ 48.01 for BART).

\begin{table}[t]
    \small
    \centering
    \setlength{\tabcolsep}{10.1pt}
    \begin{tabular}{llrr}
    \toprule
        \textbf{Model} & \textbf{Dataset} & \multicolumn{1}{c}{\textbf{\textsc{in-src}}} & \multicolumn{1}{c}{\textbf{\textsc{out-src}}} \\
        \midrule
        \multirow{3}{*}{BART} & SciTLDR & 54.53 & 0.92 \\
                              & XSum & 73.78 & 30.34 \\
                              & ScisummNet & 75.21 & 8.17 \\
        \midrule
        \multirow{3}{*}{TDSum} & SciTLDR & 56.75 & 1.37 \\
                              & XSum & 66.61 & 26.66 \\
                              & ScisummNet & 75.04 & 14.41 \\
        \bottomrule
    \end{tabular}
    \caption{ROUGE-K scores on keywords seen (\textsc{in-src}) vs.\ unseen (\textsc{out-src}) in source documents.}
    \label{ta:rouge-k-inout}
\end{table}

\paragraph{RQ2: In search of missing keywords.}
We next focus on studying the relationship between a few specific characteristics of keywords and ROUGE-K scores.
First, we look at \textit{whether models can better include keywords if they appear in source documents}.
We do this by splitting a list of pre-defined keywords into two lists, (1) an \textsc{in-src} keyword list where all the keywords appear in the source documents, (2) an \textsc{out-src} keyword list where keywords cannot be found in the source documents, and then evaluate a model with ROUGE-K using each list.
As Table \ref{ta:rouge-k-inout} shows, on both datasets and models, ROUGE-K with \textsc{out-src} keywords is notably lower than \textsc{in-src} ROUGE-K showing that when keywords are not in the source texts it is challenging for models to include them in summaries.

\begin{figure}[t]
    \centering
    \includegraphics[width=\columnwidth]{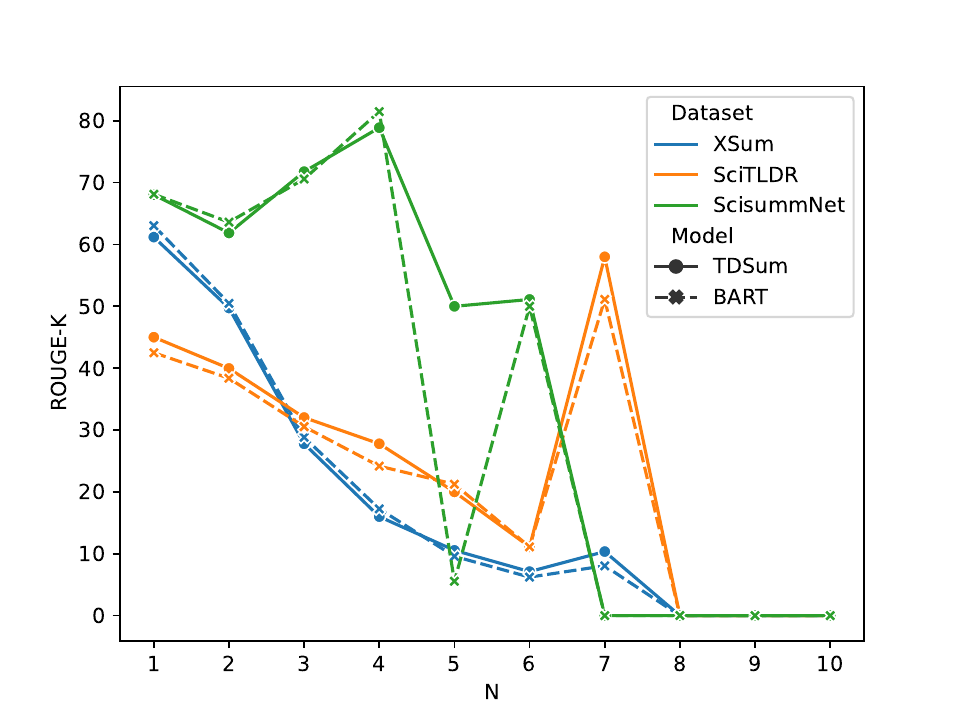}
    \caption{ROUGE-K and keyword length.}
    \label{fig:ngram-rouge-k}
\end{figure}

We next investigate whether there is a correlation between ROUGE-K and keyword length, that is, \textit{whether longer keywords are more difficult to include.}
Figure \ref{fig:ngram-rouge-k} shows that although there is one exceptional case ($N=7$), ROUGE-K scores consistently decrease as keywords become longer, indicating the difficulty of including longer keywords in summaries.
Another finding in this analysis is that while BART outperforms TDSum on XSum when keywords with all lengths are used when the n-grams are longer ($N>=5$), TDSum starts to surpass BART on ROUGE-K.

\paragraph{\newtext{LLMs on ROUGE-K}} \newtext{To shed light on the summarization behaviour of recent large language models (LLMs), we evaluate an open-source model, namely instruction fine-tuned Llama 2 \citep{touvron2023llama} in two different sizes. The prompt used in our experiments is ``Generate a summary of the following document in one sentence''. Due to our limited computational resources, we cut off inputs and outputs at 512 and 128 tokens, respectively, and also truncate all the sentences after the first one in the generated summary, if longer.
Results are shown in Table \ref{ta:llms}. While they perform remarkably well on both ROUGE and our ROUGE-K given that inferences are performed in zero-shot manner, we observe that more than half of the keywords are missing, calling for better prompting strategies. On traditional ROUGE scores, which consider the words related to writing style, Llama performs noticeably worse than fine-tuned models because fine-tuning can help models learn reference styles from the dataset. However, results are comparable on ROUGE-K. This indicates that ROUGE-K can better account for the high quality of LLMs, despite different styles between generated and reference summaries, which has been noticed when evaluating zero-shot models \citep{goyal_news_2022}.
}
\begin{table}[t]
    \small
    \centering
    \setlength{\tabcolsep}{2.8pt}
    \begin{tabular}{lccccccccc}
    \toprule
    \multirow{2}{*}{\textbf{Dataset}} & \multicolumn{4}{c}{\textbf{Llama2 7B}} & & \multicolumn{4}{c}{\textbf{Llama2 13B}} \\
      \cmidrule{2-5} \cmidrule{7-10}       & \textbf{R-1} & \textbf{R-2} & \textbf{R-L} & \textbf{R-K} & & \textbf{R-1} & \textbf{R-2} & \textbf{R-L} & \textbf{R-K} \\
      \midrule
      SciTLDR & 35.9 & 13.9 & 26.7 & 44.1 & & 36.3 & 14.6 & 27.8 & 44.4 \\
      XSum & 21.8 & 5.6 & 15.5 & 34.7 & & 22.0 & 5.6 & 15.9 & 35.9 \\
      SciNet & 46.1 & 24.1 & 32.5 & 64.6 & & 46.7 & 25.3 & 33.0 & 69.5 \\
      \bottomrule
    \end{tabular}
    \caption{\newtext{Results with Llama2 7B and 13B.}}
    \label{ta:llms}
\end{table}

\begingroup
\begin{table}[t]
\small
\centering
\begin{tabularx}{\columnwidth}{| X |}
\multicolumn{1}{c}{Generated summary (a)}\\
\hline
\textbf{Input:} Deep convolutional neural networks (CNNs) are known to be robust against label noise on extensive datasets. However, at the same time, CNNs are [...] \\
(\url{https://openreview.net/forum?id=H1xmqiAqFm})\\
\hline
\textbf{TLDR:} The authors challenge the \highlight{CNNs} \highlight{robustness} to \highlight{label noise} using ImageNet 1k tree of WordNet.\\
\hline
\textbf{BART:} We investigate the behavior of \highlight{CNNs} under class-dependently simulated \highlight{label noise}, which is generated based on the conceptual distance between classes of a large dataset.\\
\hline
\textbf{TDSum:} We show that \highlight{CNNs} are more \highlight{robust} to class-dependent \highlight{label noise} than class-independent label noise, which is generated based on the conceptual distance between classes of a large dataset.\\
\hline
\multicolumn{1}{c}{} \\
\multicolumn{1}{c}{Generated summary (b)}\\
\hline
\textbf{Input:} We explore ways of incorporating bilingual dictionaries to enable semi-supervised neural machine [...] \\
(\url{https://arxiv.org/abs/2004.02071})\\
\hline
\textbf{TLDR:} We \highlight{use bilingual dictionaries} for \highlight{data} augmentation for \highlight{neural machine translation}. \\
\hline
\textbf{BART:} We propose a simple \highlight{data} augmentation technique to address both this shortcoming.\\
\hline
\textbf{TDSum:} We propose a simple \highlight{data} augmentation technique to enable semi-supervised \highlight{neural machine translation}.\\
\hline
\end{tabularx}
\caption{Summaries generated by our models.}
\label{ta:generated-sample}
\end{table}
\endgroup

\paragraph{Model distinguishability.}
Most if not all recent summarization papers perform evaluations using multiple ROUGE metrics, yet the gap between systems is very small, making it hard to distinguish models' performance. Inspired by work from \citet{xiao_are_2022} on characterizing the distinguishability of datasets, we compute the standard deviation of scores from our models (cf.\ Section \ref{se:baseline_models}) for each metric, to see the distinguishability of ROUGE variants (larger deviation means higher distinguishability). As shown in Table \ref{ta:results}, ROUGE-K achieves the highest standard deviation among other ROUGE metrics for two extreme summarization datasets, i.e., it differentiates models better when summaries are required to be very short. We highlight this by means of the sample output shown in Table \ref{ta:generated-sample}. In (a), BART fails to include one of the keywords `robust' which is necessary to convey the purpose of the paper. In (b), the summary by BART does not mention the task the paper worked on (in this case, neural machine translation) while TDSum successfully includes it.
\section{Related work}
\label{se:related}
In the context of factual consistency evaluation \citep{kryscinski-etal-2020-evaluating,scialom-etal-2021-questeval,fabbri-etal-2022-qafacteval}, \citet{nan-etal-2021-entity} propose to use a NER model to detect hallucinated named entities in summaries. While their approach also focuses on specific words in summaries as our \mbox{ROUGE-K}, it is limited because (1) not all critical information consists of named entities, (2) strong NER models are not available in many domains, and (3) NER performance is unknown for summarization datasets.
\citet{ng-abrecht-2015-better} and \citet{zhang-etal-2020-bertscore} propose to use vector representations to compute semantic similarity between reference and candidate summaries. \citet{eyal-etal-2019-question} instead propose to use a question-answering system to assess the summary quality. While these methods can exploit semantic knowledge stored in parameters in large models, as a side-effect, they introduce `blackboxness' that hinders transparent model development. In contrast, we take a `bottom-up' approach by proposing to focus on keyword availability.
\section{Conclusion}
\label{se:conclusion}
In this paper, we proposed ROUGE-K, an extension of ROUGE to quantify how summary-relevant keywords are included in summaries. Using ROUGE-K, we showed \newstorytext{human annotators prefer summaries with more keywords} and how models often miss several essential keywords in their output.
In a variety of experiments using the baseline provided by a large pre-trained language model (BART) we showed how ROUGE-K only moderately correlates with ROUGE and BERTScore, thus indicating that it can complement them, and correlates with the length of the generated summaries on a par with ROUGE F1 and BERTScore, despite being a recall-oriented metric. Finally, we proposed four ways to guide BART to include more keywords in its summaries.
We plan in future work to further test our metric's applicability across different domains and languages, e.g., by relying on WikiLingua \citep{ladhak-etal-2020-wikilingua} and X-SciTLDR \citep{takeshita-etal-2022-xscitldr}.
\section{Limitations}
\label{se:limitations}
This work has the following limitations: (1) Our new evaluation metric, ROUGE-K, uses a heuristic to extract keywords automatically. Although it enables to obtain better and more comprehensive keywords compared to other existing methods, some nonessential words are still included thus can bring some noise into the evaluation. (2) ROUGE-K does not take the context of keywords into consideration which leaves the possibility open that generated summaries with keywords still convey the meaning of keywords wrongly. \newtext{(3) Like traditional ROUGE scores, ROUGE-K is based on hard string match, which cannot compensate for the semantics of, e.g., (near-)synonyms and paraphrases.
}



\bibliography{refs}




\end{document}